\documentclass[runningheads]{llncs}
\usepackage{graphicx}
\usepackage{comment}
\usepackage{amsmath,amssymb} 
\usepackage{color}
\usepackage{url}            
\usepackage{booktabs}       
\usepackage{amsfonts}       
\usepackage{nicefrac}       
\usepackage{microtype}      
\usepackage{graphicx}
\usepackage{mathrsfs,amsmath,amssymb,mathtools}
\usepackage{multirow}
\usepackage{bm}
\usepackage{tabularx}       
\usepackage{float}
\usepackage{array}
\usepackage[dvipsnames]{xcolor}
\usepackage{subcaption}
\usepackage{multirow}
\definecolor{caribbeangreen}{rgb}{0.0, 0.8, 0.6}

\begin{document}
	\pagestyle{headings}
	\mainmatter
	\def\ECCVSubNumber{18}  
	
	\title{Visible Feature Guidance for Crowd Pedestrian Detection}

	\titlerunning{Visible Feature Guidance for Crowd Pedestrian Detection}
	%
	\author{Zhida Huang\and
	Kaiyu Yue \and
	Jiangfan Deng \and
	Feng Zhou
	}
	\authorrunning{Z. Huang et al.}
	%
	\institute{Algorithm Research, Aibee Inc.}
	\maketitle
	
	\begin{abstract} 
	Heavy occlusion and dense gathering in crowd scene make pedestrian detection become a challenging problem, because it's difficult to guess a precise full bounding box according to the invisible human part.
	To crack this nut, we propose a mechanism called Visible Feature Guidance (VFG) for both training and inference.
	During training, we adopt visible feature to regress the simultaneous outputs of visible bounding box and full bounding box.
	Then we perform NMS only on visible bounding boxes to achieve the best fitting full box in inference.
	This manner can alleviate the incapable influence brought by NMS in crowd scene and make full bounding box more precisely.
	Furthermore, in order to ease feature association in the post application process, such as pedestrian tracking, we apply Hungarian algorithm to associate parts for a human instance.  
	Our proposed method can stably bring about 2$\sim$3\% improvements in mAP and $\text{AP}_{50}$ for both two-stage and one-stage detector.
	It's also more effective for $\text{MR}^{-2}$ especially with the stricter IoU.
	Experiments on Crowdhuman, Cityperson, Caltech and KITTI datasets show that visible feature guidance can help detector achieve promisingly better performances.
	Moreover, parts association produces a strong benchmark on Crowdhuman for the vision community.
	\end{abstract}
	\section{Introduction}
	\begin{figure*}[t]
		\centering
		\scriptsize
		\includegraphics[width=1.0\textwidth]{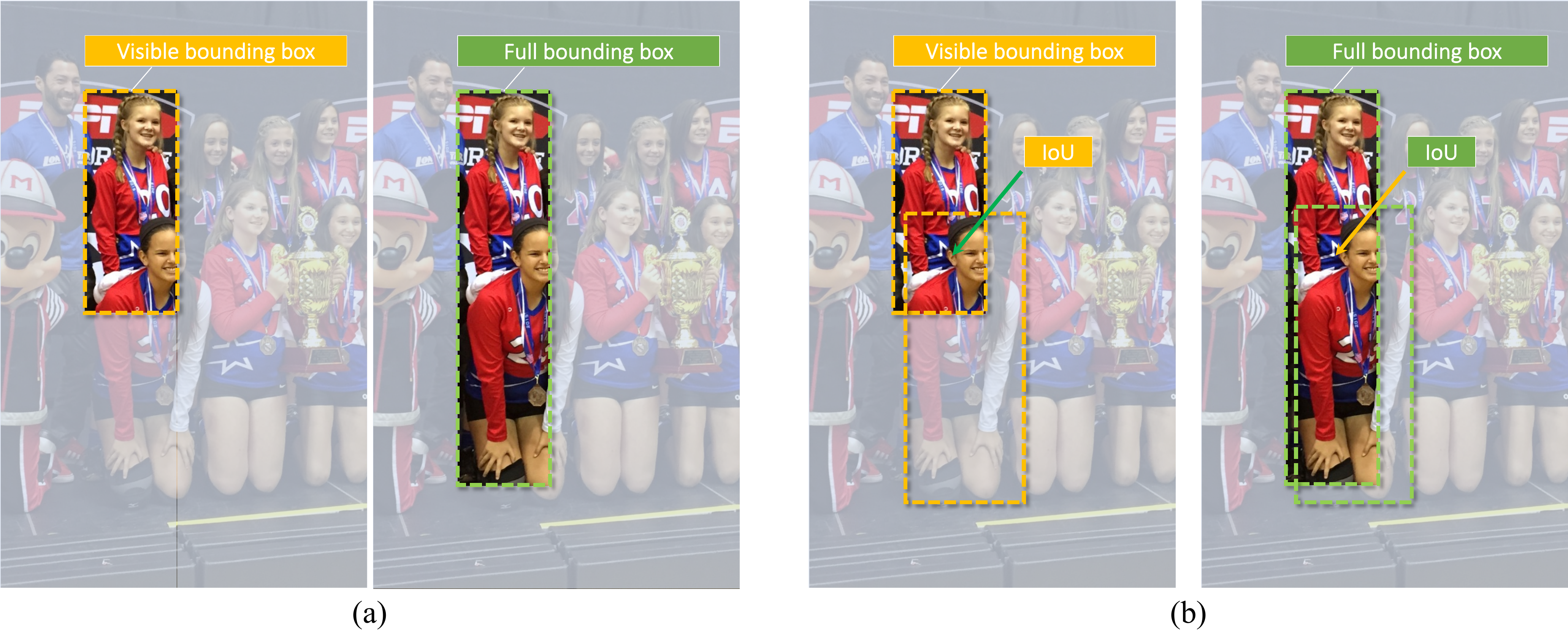}
		\caption{
			\small{
				\textbf{Visible bounding box \textit{vs.} Full bounding box.}
				(a) We illustrate the visible and full bounding box of a same human instance in a crowd scene.
				(b) Conspicuously, the full bounding box has a dramatically larger overlap (IoU) than visible bounding box.
		}}
		\label{fig:intro}
		\vspace{-1.5\baselineskip}
	\end{figure*}
	Pedestrian detection is widely applied in a number of tasks, such as autonomous driving, robot-navigation and video surveillance. 
	Many previous research efforts~\cite{ouyang2012discriminative,zhang2015filtered,zhang2016faster,bodla2017soft,wang2018repulsion,zhang2018occluded,zhou2018bi,pang2019mask} have been made to improve its performance.
	Although reasonably good performance has been achieved on some benchmark datasets for detecting non-occluded, slightly occluded and crowd pedestrians, it's still far from being satisfactory for detecting heavily occluded and crowded pedestrians.
	To push the cutting-edge boundary, this work aims to propose a novel and simple method to precisely detect pedestrians in the heavily crowd scene. 
	
	In detection, bounding box (or box) is commonly used to represent the object location in an image. 
	According to its application scenario, it generally has two distinct types: visible one and full one.
	As shown in Fig.~\ref{fig:intro}(a), a visible box only covers the visible part of an object.
	In contrast, the full box covers the whole expected region of an object even though it's occluded by the other objects.
	However, the full bounding box needs detector to estimate the invisible border. 
	This property hinders the detector to produce highly precise outputs.
	We analyze this problem in two folds:
	(1) First, intuitively, it's unreasonable to use invisible feature to guide a regressor for producing the full bounding box.
	This will easily plunge the detector into a random process to guess the border for an occluded object.
	(2) Second, for two heavily crowed and occluded instances, it's hard to perform Non-Maximum-Suppression (NMS) well on their full boxes to separate them precisely, because there is a large overlap between the full bounding boxes of them, which would be easily merged into a same box by NMS, as shown in Fig.~\ref{fig:intro}(b).
	%
	%
	
	To overcome these two obstacles, we propose an effective mechanism named Visible Feature Guidance (VFG).
	For training, we restrict RPN only produces the visible bounding-box proposals. Then we adopt the visible feature to simultaneously regress the visible bounding box and full bounding box.
	This is motivated by the visible part of object that represents the clear appearance to pinpoint the area for guiding the regressor where should be focused on.
	In inference, we use the visible feature as well to generate the coupled pair of the visible box and full box.
	But we only perform NMS on visible bounding-boxes to shave the confused ones, then achieving the corresponding full bounding box in the end.
	The insight of this behavior is that we found the overlap of visible bounding boxes between heavily crowed instances is smaller than that of full bounding boxes, as illustrated in Fig.~\ref{fig:intro}. 
	Using visible boxes to perform NMS can effectively prevent the boxes from being merged into a same box for the common case of two heavily crowed instances (e.g., the IoU of full boxes is larger than threshold, but the IoU of visible boxes is smaller than threshold).
	%
	
	In addition, some application scenario not only requires the simple output of the whole object bounding box, also requires the precise localization for its part (e.g., head, face).
	They need detector to stably generate the coupled localization of the main object and its semantic parts.
	In this paper, we transform the parts association task into an assignment problem solved by Hungarian algorithm \cite{kuhn1955hungarian}. 
	This simple post process can effectively produce the combined output of full bounding box, visible bounding box and part localizations for a same instance.
	Experiments on Crowdhuman \cite{shao2018crowdhuman} introduce a strong baseline and benchmark for the parts association task.
	
	Overall, our main contributions in this work are summarized into three folds:
	\\
	- We present visible feature guidance (VFG) mechanism.
	It regresses visible bounding box and full bounding box simultaneously during training.
	Then we perform NMS on visible bounding-boxes to achieve the best fitting full box in inference in order to mitigate the influence of NMS on full boxes.
	\\
	- We transform parts association into a linear assignment problem, including body-head, body-face or head-face association. 
	We build a strong benchmark on Crowdhuman \cite{shao2018crowdhuman} dataset for this parts association task.
	\\
	- Experiments are carried out on four challenging pedestrian and car detection datasets, including KITTI \cite{geiger2012we}, Caltech \cite{dollar2009pedestrian}, CityPerson \cite{zhang2017citypersons}, and Crowdhuman \cite{shao2018crowdhuman}, to demonstrate that our proposed method works stably better than previous methods.
	
	\section{Related Work}
	
	With the development of deep convolution neural network (CNN), object detection has made a dramatic improvement in both performance and efficiency. 
	Recently, detection methods can be roughly divided into two types: two-stage and one-stage. 
	Two-stage detector, such as Faster R-CNN \cite{ren2015faster}, R-FCN \cite{Dai2016R} and Mask R-CNN \cite{he2017mask} first generates the region proposals and then refine these coarse boxes after an scale-invariant feature aggregating operation named ROI-Pool \cite{girshick2015fast} or RoI-Align \cite{he2017mask}. 
	This coarse-to-fine process leads to achievements of top performance. 
	One-stage detector, such as YOLO \cite{redmon2016you}, SSD \cite{liu2016ssd} and RetinaNet \cite{lin2017focal}, predicts locations directly in an unified one-shot structure which is fast and efficient. Recently, many novel anchor-free detectors have emerged \cite{LawCornerNet,DuanCenterNet,TianFCOS,KongFoveaBox}, these methods cancel the hand-craft tuning of pre-defined anchors, and simplify the training process. 
	In the area of pedestrian detection, previous works such as such as Adapted Faster R-CNN \cite{zhang2017citypersons}, \cite{liu2019high} design an anchor-free method to predict the body center and scale of instance box.

	However, crowd occlusion is sitll one of the most important and difficult problems in pedestrian detection. 
	Repulsion loss \cite{wang2018repulsion} is designed to penalize the predicted box to avoid shifting to the wrong objects and push it far from the other ground-truth targets.
	Bi-box regression \cite{zhou2018bi} aims to use two branches for regressing the full and visible boxes simultaneously.
	Occlusion-Aware RCNN \cite{zhang2018occlusion} proposes an aggregation loss to enforce proposals closely for objects.
	It also uses the occlusion-aware region of interest (RoI) pooling with structure information.
	Compared with \cite{zhou2018bi} and \cite{zhang2018occlusion}, our proposed method is more concise with using visible box to produce full box directly.
	
	Non-Maximum-Suppression (NMS) is mostly used by object detection in post-processing stage. 
	Firstly, it sorts all detection boxes based on their scores. 
	Then the detection box with maximum score is selected and all other detection boxes which have large overlap with the selected one are suppressed. 
	This process is running recursively until there are no remaining boxes. 
	Soft-NMS \cite{bodla2017soft} decreases detection scores using a continuous function of their overlaps such as Guassian Kernel.
	It drops the candidate boxes progressively and carefully.
	Adaptive-NMS \cite{liu2019adaptive} designs a detector sub-network that learns the target density to decide what dynamic suppression threshold should be applied to an instance.  
	
	However, NMS will remove valid boxes in crowded scene.
	Soft-NMS would remove valid boxes and introduce false positive boxes.
	Adaptive-NMS need a additional network to generate the threshold value.
	In this work, we propose VFG-NMS which simply performs NMS with visible boxes and achieves the corresponding full boxes according to the indexes of remaining visible boxes.
	This manner make detection more robust and less sensitive to NMS threshold. There is a contemporaneous idea published in R2NMS\cite{2020NMS}.  

	\section{Methodology}
\begin{figure*}[h]
		\centering
		\begin{minipage}[t]{0.5\linewidth}
			\centering
			\includegraphics[width=1.0\textwidth]{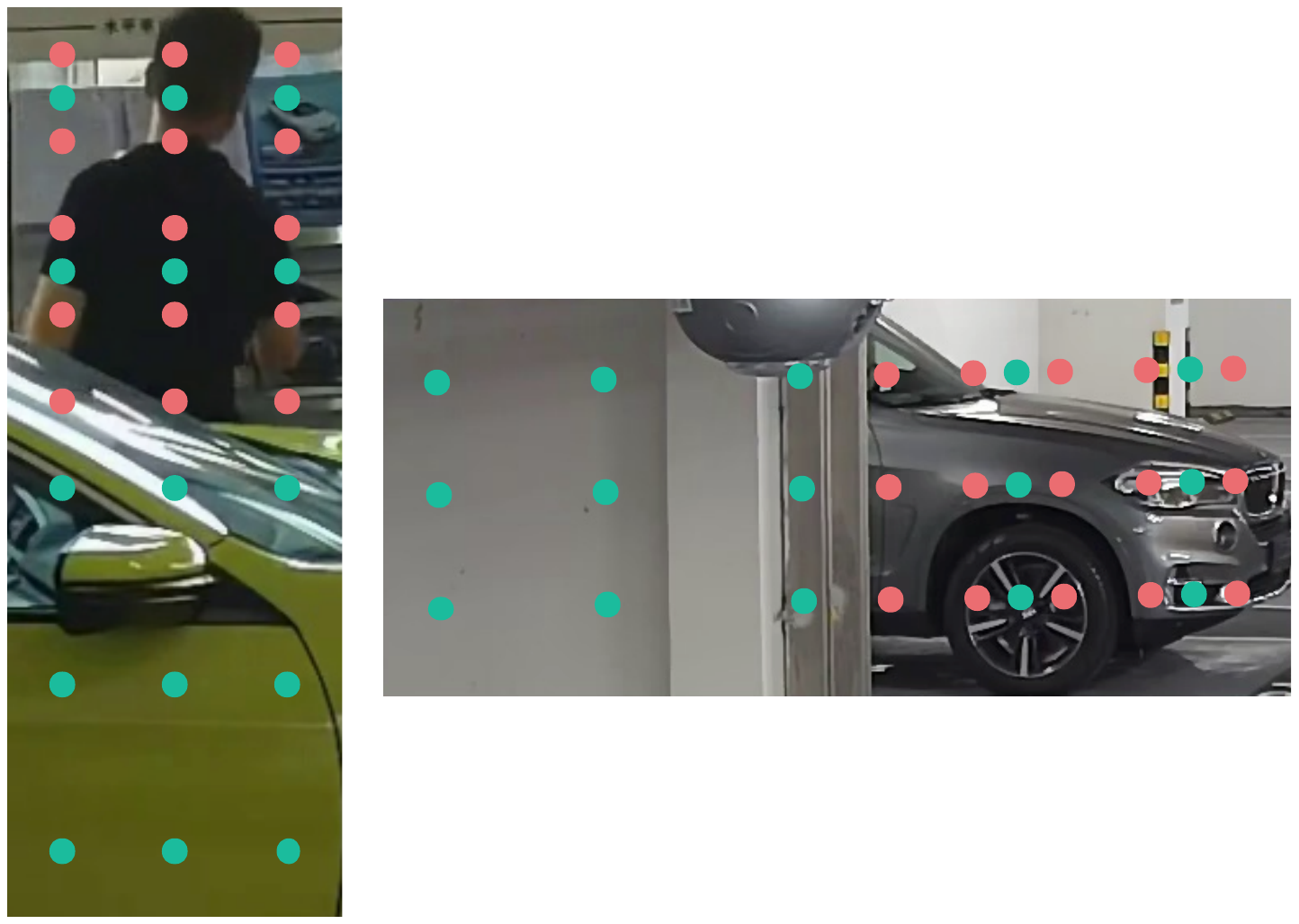}
			\caption{\small{\textbf{Feature regions}
					produced by visible (red) and full boxes (green).}}
			\label{fig:roi_region}
		\end{minipage}
		\begin{minipage}[t]{0.48\linewidth}
			\centering
			\includegraphics[width=1.0\textwidth]{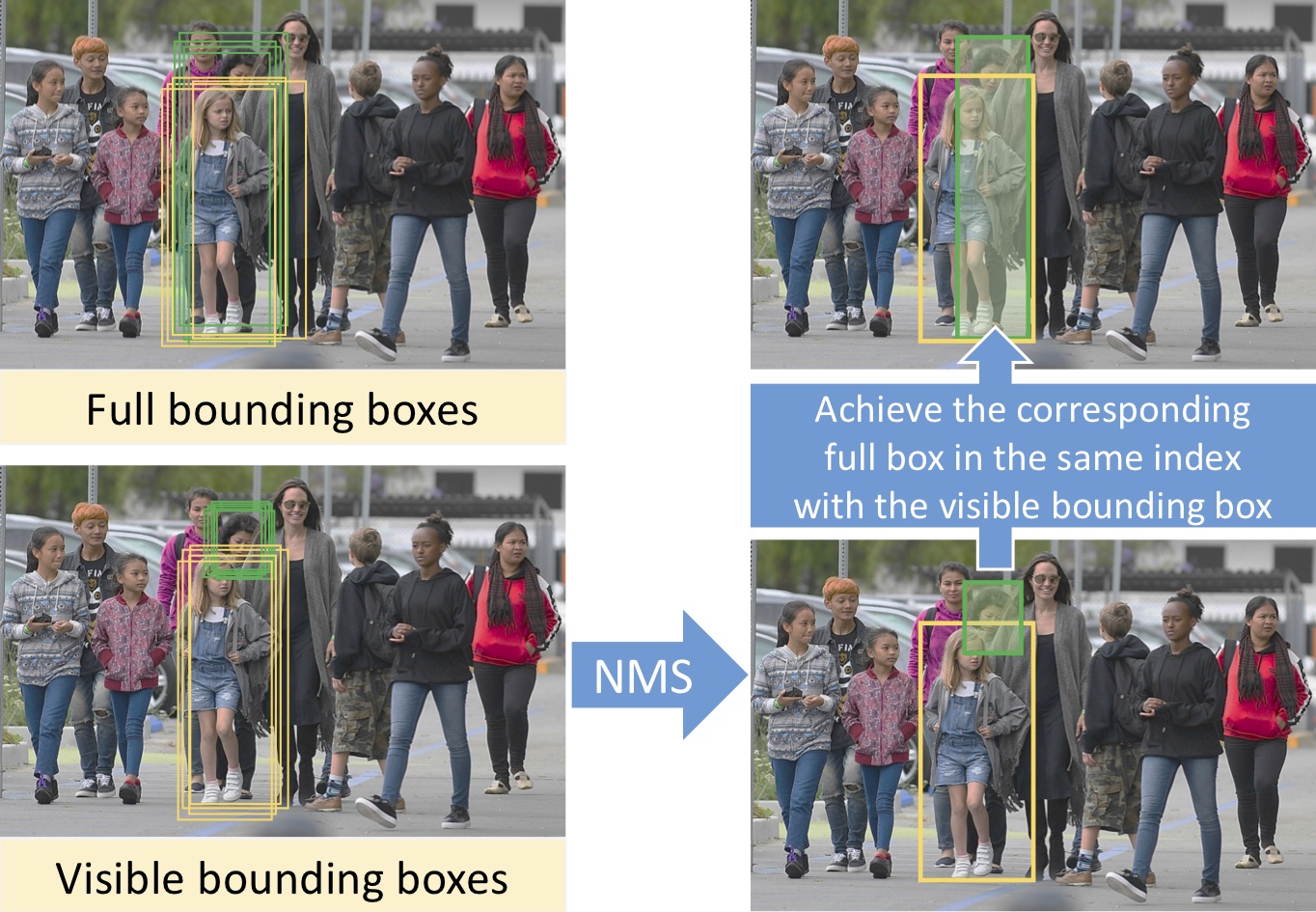}
			\caption{\textbf{VFG-NMS} is done with visible boxes.}
			\label{fig:nms_visible}
		\end{minipage}
		\vspace{-1.5\baselineskip}
	\end{figure*}

	Our VFG method is general and independent of detector types.
	Considering the existence of two typical frameworks: two-stage and one-stage, we go into the details of building the VFG module with these two models respectively. 
	Then we discuss the method of parts association, which is the downstream process of pedestrian detection.
	\begin{figure}[t]
		\centering
		\includegraphics[width=1.0\textwidth]{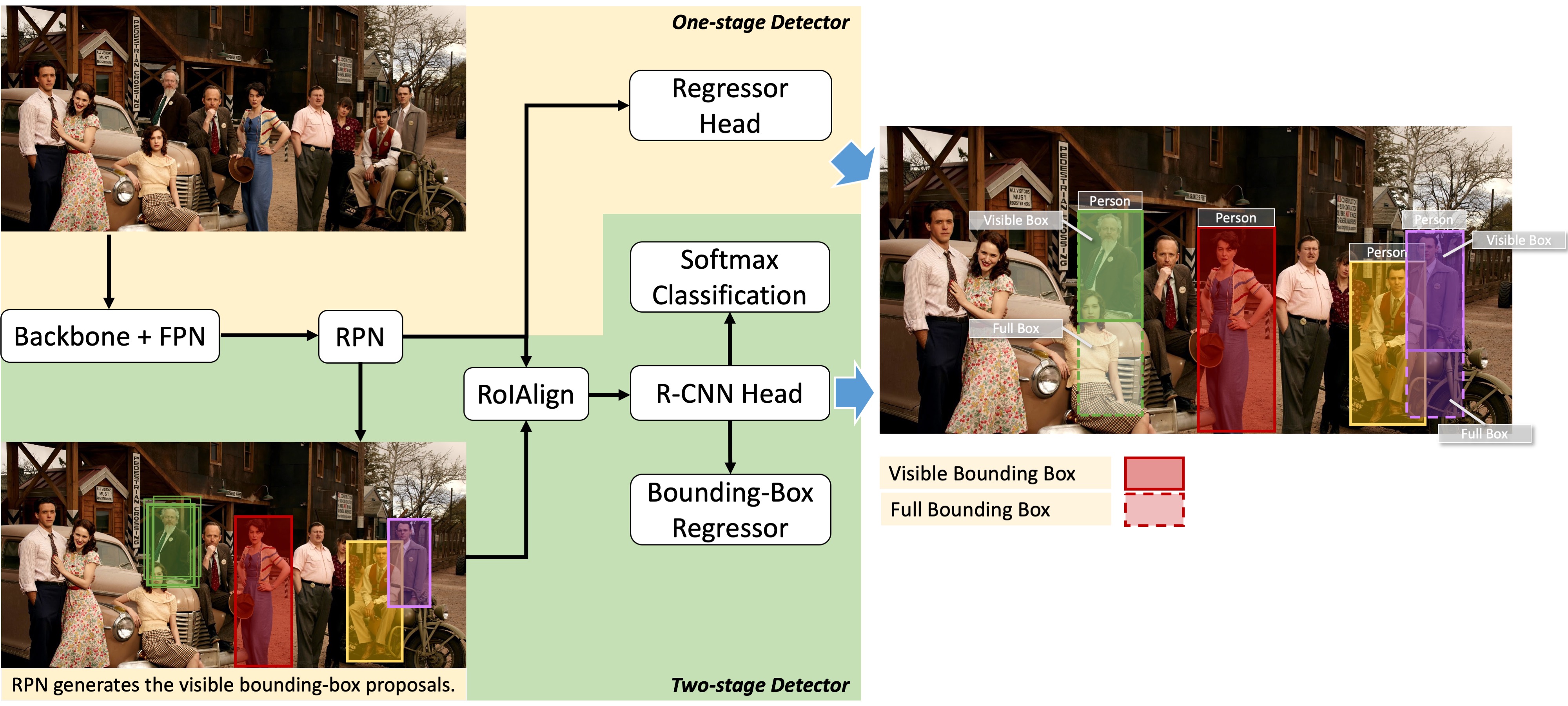}
		\caption{\small{
				\textbf{Visible feature guidance (VFG) for one-stage and two-stage detector.} 
				We let RPN to regress visible box and RCNN to regress pairs of visible and full boxes simultaneously by using visible feature for two-stage detector. 
				For one-stage detector, regression head directly regresses the pairs of visible and full boxes.
		}}
		\label{fig:net_arch}
	\end{figure}
	\\[5pt]
	\textbf{Architecture.} The proposed method can be easily implemented into current detectors.
	Fig.~\ref{fig:net_arch} shows the main structure of VFG method. 
	An input image is firstly filled into backbone and FPN (if exists) to extract features.
	Then, for two-stage detector like Faster R-CNN~\cite{ren2015faster}, the Region Proposal Network (RPN)~\cite{ren2015faster} generates visible bounding-box proposals and sends them into RoIPool~\cite{girshick2015fast} or RoIAlign~\cite{he2017mask}, from which the RoI feature is aggregated.
	After that, the R-CNN head uses these visible RoI features to regress pairs of visible and full boxes in the \textit{parallel} manner. 
	For one-stage detector, following similar scheme, visible bounding boxes are generated as base proposals, which are used by the final submodule to regress visible and full bounding boxes at the same time. Compared to the typical pipeline of detector, this structure includes two key differences: 
	(1) The detector regresses eight coordinate values for both visible and full bounding box, instead of the conventional four values.
	(2) The positive and negative sampling during training is only based on the IoU of visible proposals and ground truth.
	\\[5pt]
	\textbf{Visible Feature Guidance.} As shown in Fig.~\ref{fig:roi_region}, {\color{red}{red}} and {\color{caribbeangreen}{green}} points represent the feature region of visible and full box of an object (a pedestrian or a car in the image) respectively.
	For the two-stage detector, RoIAlign could be applied with one of two potential feature regions: visible region (in red points) and full region (in green points).
	Previous methods in default choose full feature regions to perform RoIAlign for guiding regression.
	However, the features of full box is half occupied by the background, which is harmful that plunges detector into a random process to regress the invisible border.
	Fortunately, the feature region of visible box focus on the valid part of an object, which has a better potential to estimate entire structure precisely.
	Thus we utilize visible feature region to apply RoIAlign to produce a clear and effective feature represents for guiding the bounding-box regressor and classifier. 
	For one-stage detector, in the same sense, taking visible box to regress full box can be considered as adding attention factors directly to the visible region, facilitating and guiding the network to learn the object shape and estimate the full box location.
	\\[5pt]
	\textbf{Multi-task Loss.}
	In the procedure of training, we propose multi-task training to supervise and guide bounding box regressor and classifier using visible feature regions.
	The bounding-box regressor produces two outputs for a instance in parallel: visible box and full box. 
	Let the center coordination of visible bounding-box proposal for an object\footnote{
		Here all the notations ignore the object classes.
		In default, the notation indicates the representation for a instance of one object class. 
	} be $(x_{v}^p, y_{v}^p)$ within width $P_w$ and height $P_h$. 
	The ground-truth of visible box is $(x_{v}^g, y_{v}^g, V_w, V_h)$ and which of full box is $(x_{f}^g, y_{f}^g, F_w, F_h)$. 
	So the visible and full box regression formula is:
	\begin{align}
	\triangle_{x_{v}}^* & = \frac{x_{v}^g - x_{v}^p}{P_w},  \triangle_{y_{v}}^* = \frac{y_{v}^g - y_{v}^p}{P_h}, 
	\triangle_{w_{v}}^* = \log \frac{V_w}{P_w},  \triangle_{h_{v}}^* = \log \frac{V_h}{P_h} \\
	\triangle_{x_{f}}^* & = \frac{x_{f}^g - x_{f}^p}{P_w},  \triangle_{y_{f}}^* = \frac{y_{f}^g - y_{f}^p}{P_h}, 
	\triangle_{w_{f}}^*  = \log \frac{F_w}{P_w},  \triangle_{h_{f}}^* = \log \frac{F_h}{P_h} 
	\end{align}
	The multi-task loss $L$ on each labeled RoI to jointly train for classification and bounding-box regression, including visible box and full box:
	\begin{align}
	L = L_{cls}(c,c^*) + \lambda_{loc}L_{loc}(t, t^*)
	\end{align}
	where $c$ is predicted output and $c^*$ is ground-truth.  
	$L_{cls}$ denotes the loss for classification: commonly softmax cross-entropy loss for two-stage detector and focal loss \cite{lin2017focal} for one-stage detector. 
	$t = (\triangle_{x_{v}}, \triangle_{y_{v}},   \triangle_{w_{v}}, \triangle_{h_{v}} ,  \triangle_{x_{f}}, \triangle_{y_{f}}, \triangle_{w_{f}}, \triangle_{h_{f}} )$ and $t^* = (\triangle_{x_{v}}^*, \triangle_{y_{v}}^*,   \triangle_{w_{v}}^*, \triangle_{h_{v}}^* ,\triangle_{x_{f}}^*, \triangle_{y_{f}}^*, \triangle_{w_{f}}^*, \triangle_{h_{f}}^* )$ represent eight predicted and ground-truth parameterized targets.  
	$L_{loc}$ is the loss function for regression, such as smooth-$L_1$ loss or IoU loss \cite{yu2016unitbox}. $\lambda_{loc}$ is a loss-balancing parameter and we set $\lambda_{loc} = 3$.
	\\[5pt]
	\textbf{VFG-NMS.} In the procedure of inference, NMS is a common post-processing method for object detection. 
	However, it may remove valid bounding boxes improperly in the heavily crowded scene.
	To address this issue, we propose VFG-NMS: using the less crowded visible boxes to guide full boxes filtering. In Fig.~\ref{fig:nms_visible}, 
	the detector outputs visible and full boxes in pairs, we send all visible boxes into a standard NMS and get the corresponding full boxes according to remaining indices.
	Through this operation, instances which are highly occluded in full box but less occluded in visible box can be preserved.
	Compared to Soft-NMS \cite{bodla2017soft} and adaptive-NMS \cite{liu2019adaptive}, VFG-NMS alleviates the burden of tuning hyper-parameters, and is more efficient. 
	\\[5pt]
	\textbf{Parts Association.} After achieving the precise full bounding box using VFG, we are still need more precise localization of the semantic parts to be associated with full box in some scenarios, such as pedestrian re-identification and tracking.
	Because their performances can be promoted by the fine-grained feature aggregation from the parts association (e.g., body-head, body-face and head-face association).
	To reach this goal, the pioneer works use greedy algorithm to couple the part bounding boxes.
	But this behavior would recall a number of incorrect associations, particularly in the crowd scene. 
	In this paper, we pose the body-parts association problem as a Linear Assignment problem. 
	We resolve it by the efficient Hungarian algorithm~\cite{kuhn1955hungarian} to minimize the assignment cost.
	The cost matrix is constructed by one of two measurements\footnote{
		Generally, for the association of which body-part is in a small overlap, e.g. body-head association, we select one of distances to build the cost matrix.
		Vice in versa, for the association which the overlap is too large, e.g. head-face association, we prefer to use the IoU metric to constrain the parts' boxes.
	}: geometry distance and IoU metric.
	%
	With $R$ bodies and $C$ heads overall the instances in an image, we can use appropriate distance metric and IoU as the constrain condition to construct  an $N_R \times N_C$ ($N_R \leq N_C$) cost matrix $D$ according to the spatial relations of body and part, which are belonging to the same human instance.
	The standard assignment problem can be optimized by minimizing the cost function:
	\begin{align}
	\label{eq:la}
	X^* &= \arg \min \sum_{i=1}^{N_R} \sum_{j=1}^{N_C} \rho_{i,j} x_{i,j} ; \ st.  \sum_{j=1}^{N_C} x_{i,j} = 1, \forall i; \   \sum_{i=1}^{N_R} \leq 1, \forall j;   x_{i,j} \in \{0, 1\}
	\end{align}
	where $\rho_{i,j}$ is the cost function to measure the relation of $i-$th body and $j-$th part, which can be distance function (e.g. Euclidean distance) or IoU metric. 
	After solving the optimization problem of unknown $x_{ij}$, we can get body-head association. 

	\section{Experiment}
	We have done with massive experiments on four datasets: CityPerson \cite{zhang2017citypersons}, Crowdhuman \cite{shao2018crowdhuman}, Caltech \cite{dollar2009pedestrian} for pedestrian detection and KITTI \cite{geiger2012we} for car detection. 
	All of three pedestrian datasets provide visible and full bounding-box annotations.
	Specially, Crowdhuman \cite{shao2018crowdhuman} has extra head bounding-box annotations for human instances. 
	Since KITTI only provides full boxes, we use Mask R-CNN pre-trained on COCO to generate the visible boxes of car and associate it with labeled full boxes by using Hungarian Algorithm.
	
	\subsection{Dataset and Evaluation Metrics}
	The CityPerson \cite{zhang2017citypersons} dataset is originated from the semantic segmentation dataset Cityscape \cite{cordts2016cityscapes} for crowed pedestrian detection. 
	Visible bounding boxes are auto-generated from instance segmentation of each pedestrian.
	The Crowdhuman \cite{shao2018crowdhuman} dataset is targeted for addressing the task of crowd pedestrian detection. 
	It provides complete annotations: full bounding box, visible box, and head box.
	Furthermore, compared to other human detection datasets, it's a larger-scale dataset with much higher crowdness.
	Caltech \cite{dollar2009pedestrian} is one of several predominant datasets for pedestrian detection and Zhang \textit{et al.} \cite{zhang2016far} provided refined full box annotations. 
	KITTI \cite{geiger2012we} is the dataset for autonomous driving scene and it only provides full box annotations for the classes of pedestrian, car, tram, truck and van.
	
	In our experiments, we follow the standard Caltech evaluation\cite{dollar2009pedestrian} metrics. 
	The log miss-rate averaged over FPPI (false positive per-image) range of $[10^{-2}, 10^0]$ (denoted as $\text{MR}^{-2}$) is used to evaluate the pedestrian detection performance (lower is better). 
	We also use the standard evaluation of object detection, including mAP, recall and $\text{AP}_{50}$ to investigate the performance of pedestrian detection. 
	For car detection on KITTI, unlike previous works follow PASCAL \cite{everingham2010pascal} criteria,
	we use COCO evaluation criteria with the stricter IoU to fully report performances.
	
	\subsection{Implementation details} 
	We implement our VFG method in Faster R-CNN \cite{ren2015faster} (FRCNN) and RetinaNet \cite{lin2017focal} with FPN \cite{lin2017feature} structure. ResNet-50 \cite{he2016deep} pre-trained on ImageNet \cite{deng2009imagenet} is adopted as backbone. We set the anchor ratios for full box into \{1.0, 1.5, 2.0, 2.5, 3.0\} and our VFG method of visible box into \{0.5, 1.0, 2.0\}. During training, we use a total batch size of 8 and 16 for Faster R-CNN and RetinaNet respectively. Stochastic Gradient Descent (SGD) solver is used to optimize the networks on 8 1080Ti GPUs. We set the weight decay to 0.0001 and momentum to 0.9. The threshold of greedy-NMS and soft-NMS with the linear method is 0.5.
	
	When training on Crowdhuman, images are resized so that the short side is 800 pixels while the long side does not exceed 1400 pixels.
	For Faster R-CNN, models are trained for 72k iters in total, the initial learning rate is set to 0.02 and decreased by a factor of 10 for 48k and 64k iters.
	For RetinaNet, models are trained for 36k iters in total, the initial learning rate is set to 0.01 and decreased by a factor of 10 for 24k and 32k iters.
	Multi-scale training/testing are not applied to ensure fair comparisons.
	On Cityperson, we train Faster R-CNN for 9k iters in total using the base learning rate of 0.02 and decrease it by a factor of 10 after 6k and 8k iters.
	On KITTI, we train Faster R-CNN for 18k iters in total using the base learning rate of 0.02 and decrease it by a factor of 10 after 12k and 16k iters. 
	Original input image size was used to evaluate on Cityperson and KITTI.
	
	\subsection{Experimental Results}
	\subsubsection{Crowdhuman Dataset.}
	We evaluate Faster R-CNN and RetinaNet with VFG on Crowdhuman. 
	Table.~\ref{table:cfbp} shows that, in Faster R-CNN, our proposed VFG method outperforms the baseline, bringing about 3.5\% and 4.5\% improvement in mAP and recall while reducing $\text{MR}^{-2}$ by 0.84\% and 0.74\% in {\fontfamily{qcr}\selectfont Reasonable} and {\fontfamily{qcr}\selectfont All} sets respectively. As comparison, soft-NMS can improve mAP obviously but showing little effect in $\text{MR}^{-2}$. 
	In the case of RetinaNet, soft-NMS decreases performance a lot for introducing false positives in crowded scene. On the contrary, our VFG method can still bring improvement in mAP.
	
	
	\begin{table}
		\centering
		\scriptsize
		\caption{
			\small{\textbf{Performances of full bounding box} on Crowdhuman}
		}
		\label{table:cfbp}
		\begin{tabular}{llllllll}
			\cmidrule{1-8}
			\multirow{2}{*}{method} & \multirow{2}{*}{mAP} & \multirow{2}{*}{$\text{AP}_{50}^{\text{bbox}}$}  & \multirow{2}{*}{Recall}  & \multicolumn{4}{c}{$\text{MR}^{-2}$}  \\ \cline{5-8} & & & & {\fontfamily{qcr}\selectfont Reasonable} & {\fontfamily{qcr}\selectfont Small} & {\fontfamily{qcr}\selectfont Heavy} & {\fontfamily{qcr}\selectfont All} \\
			\toprule
			FRCNN+greedy-NMS(offical)  & -- & 84.95 & 90.24   & -- & -- & -- & 50.42 \\
			FRCNN+adaptive-nms(offical)  & -- & 84.71 & 91.27   & -- & -- & -- & 49.73 \\
			\cmidrule{1-8}
			FRCNN+greedy-NMS  & 47.9 & 83.1  & 87.8  & 25.49 & 25.29 & 46.76 & 49.02 \\
			FRCNN+soft-NMS  & 50.8 & 85.8  & \textbf{92.8}  & 25.49 & 24.98 & \textbf{46.65} & 49.02 \\
			FRCNN+VFG w/o VFG-NMS  & 49.5 & 84.5  & 89.3  & 24.95 & 24.94 & 47.97 & 48.33 \\
			FRCNN+VFG & \textbf{51.4}  & \textbf{86.4} & 92.3  & \textbf{24.65} & \textbf{24.95} & 47.25 & \textbf{48.28} \\
			\toprule
			RetinaNet+greedy-NMS(official)  & -- & 80.83 & 93.28   & -- & -- & -- & 63.33 \\
			RFBNet+adaptive-nms(official)  & -- & 79.67 & 94.77   & -- & -- & -- & 63.03 \\
			\cmidrule{1-8}
			RetinaNet+greedy-NMS  & 42.1 & 77.7 & 85.1  & 33.70 & 35.45 & \textbf{52.85}  & \textbf{57.90} \\
			RetinaNet+soft-NMS  & 40.2 & 70.1  & 87.8  & 58.0 & 49.37 &  67.76 &  77.37 \\
			RetinaNet+VFG & \textbf{47.0}  & \textbf{82.3} & \textbf{91.0} & \textbf{33.41} & \textbf{32.08} & 53.71 & 58.13 \\
			\cmidrule{1-8}
		\end{tabular}
		\vspace{-1.5\baselineskip}
	\end{table}
	
	\subsubsection{Cityperson Dataset.}
	Experiments on Cityperson follow the evaluation standard in RepLoss \cite{wang2018repulsion} and OR-CNN \cite{zhang2018occlusion}, in which the {\fontfamily{qcr}\selectfont Reasonable} part (occlusion $<$ 35\%) of the validation set is divided into {\fontfamily{qcr}\selectfont Partial} (10\% $<$ occlusion $<$ 35\%) and {\fontfamily{qcr}\selectfont Bare} (occlusion $ \leq $ 10\%) subsets.
	
	In Table. \ref{table:cpbp}, our VFG method has obvious better performance compared to baseline, improving the mAP by 2.1\% and reduce the $\text{MR}^{-2}$ by 1.8\%. Compared to Soft-NMS, VFG has competitive performances in mAP and better results in $\text{MR}^{-2}$. Especially, when we set the confidence threshold to 0.3, which is often applied in real usage, our VFG method suggests strong superiority.
	\begin{table}[t]
		\centering
		\scriptsize
		\caption{\small{\textbf{Performances of full bounding box} on Cityperson.}}
		\label{table:cpbp}
		\begin{tabular}{lclclclclclclclc}
			\cmidrule{1-9}
			\multirow{2}{*}{method} & \multirow{2}{*}{score} & \multirow{2}{*}{mAP} & \multirow{2}{*}{$\text{AP}_{50}^{\text{bbox}}$}  & \multirow{2}{*}{Recall}  & \multicolumn{4}{c}{$\text{MR}^{-2}$}  \\ \cline{6-9} & & & & & {\fontfamily{qcr}\selectfont Reasonable} & {\fontfamily{qcr}\selectfont Heavy} & {\fontfamily{qcr}\selectfont Partial} & {\fontfamily{qcr}\selectfont Bare} \\
			\toprule
			FRCNN+greedy-NMS  & 0.05 & 52.8 & 80.6 & 84.5   & 12.87 & 50.99 & 13.15 &  7.73\\
			FRCNN+soft-NMS   & 0.05 & \textbf{54.9} & \textbf{82.9} & \textbf{89.1}  & 12.76 & \textbf{50.91} & 13.14 & 7.42 \\
			FRCNN+VFG w/o VFG-NMS  & 0.05 & 53.8 & 81.0 & 84.6   & 11.63 & 51.41 & 12.36 &  6.95\\
			FRCNN+VFG & 0.05 & 54.8 & 82.3 & 86.9   & \textbf{11.04} & 50.94 & \textbf{11.56} & \textbf{6.69} \\
			\cmidrule{1-9}
			FRCNN+greedy-NMS  & 0.3 & 51.4 & 77.3 & 79.7   & 12.88 & 51.69 & 13.44 & 7.82 \\
			FRCNN+soft-NMS   & 0.3 & 51.9 & 78.1 & 80.3  & 12.76 & 51.61 & 13.40 & 7.59 \\
			FRCNN+VFG w/o VFG-NMS  & 0.3 & 52.5 & 77.6 & 79.8   & 11.67 & 52.01 & 12.47 &  6.98\\
			FRCNN+VFG & 0.3 & \textbf{53.4} & \textbf{79.1} & \textbf{81.2}   & \textbf{11.04} & \textbf{51.46} & \textbf{11.68} & \textbf{6.73} \\
			\cmidrule{1-9}
		\end{tabular}
		\vspace{-1.5\baselineskip}
	\end{table}
	\begin{table}[t]
		\centering
		\scriptsize
		\caption{\small{
				\textbf{Comparison with state-of-the-art methods} on Cityperson. 
				All the experiments are done with the original input size.
		}}
		\label{table:ccp}
		\begin{tabular}{llcccc}
			\cmidrule{1-6}
			method & Backbone & {\fontfamily{qcr}\selectfont Reasonable} & {\fontfamily{qcr}\selectfont Heavy} & {\fontfamily{qcr}\selectfont Partial} & {\fontfamily{qcr}\selectfont Bare} \\
			\toprule
			Adapted-FRCNN \cite{zhang2017citypersons} & VGG-16 & 15.40 & -- & -- & -- \\
			FRCNN Adaptive-NMS \cite{liu2019adaptive} & VGG-16 & 12.90 & 56.40  & 14.40 & 7.00   \\
			OR-CNN \cite{zhang2018occlusion} & VGG-16 & 12.80 & 55.70  & 15.30 & 6.70   \\
			RepLoss \cite{wang2018repulsion} & ResNet-50 & 13.20 & 56.90  & 16.80 & 7.60   \\
			TLL \cite{song2018small} & ResNet-50 & 15.5 & 53.60  & 17.20 & 10.0  \\
			ALFNet \cite{liu2018learning} & ResNet-50 & 12.00 & 51.90  & \textbf{11.4} & 8.40   \\
			FRCNN+greedy & ResNet-50 & 12.87 & 50.99 & 13.15 &  7.73   \\
			FRCNN+VFG & ResNet-50 & \textbf{11.04} & \textbf{50.94} & 11.56 & \textbf{6.69}   \\
			\cmidrule{1-6}
		\end{tabular}
		\vspace{-1.5\baselineskip}
	\end{table}
	\begin{figure*}[tbp]
		\centering
		\begin{subfigure}[t]{0.315\textwidth}
			\captionsetup{font={scriptsize}} 
			\centering
			\includegraphics[width=1.0\textwidth]{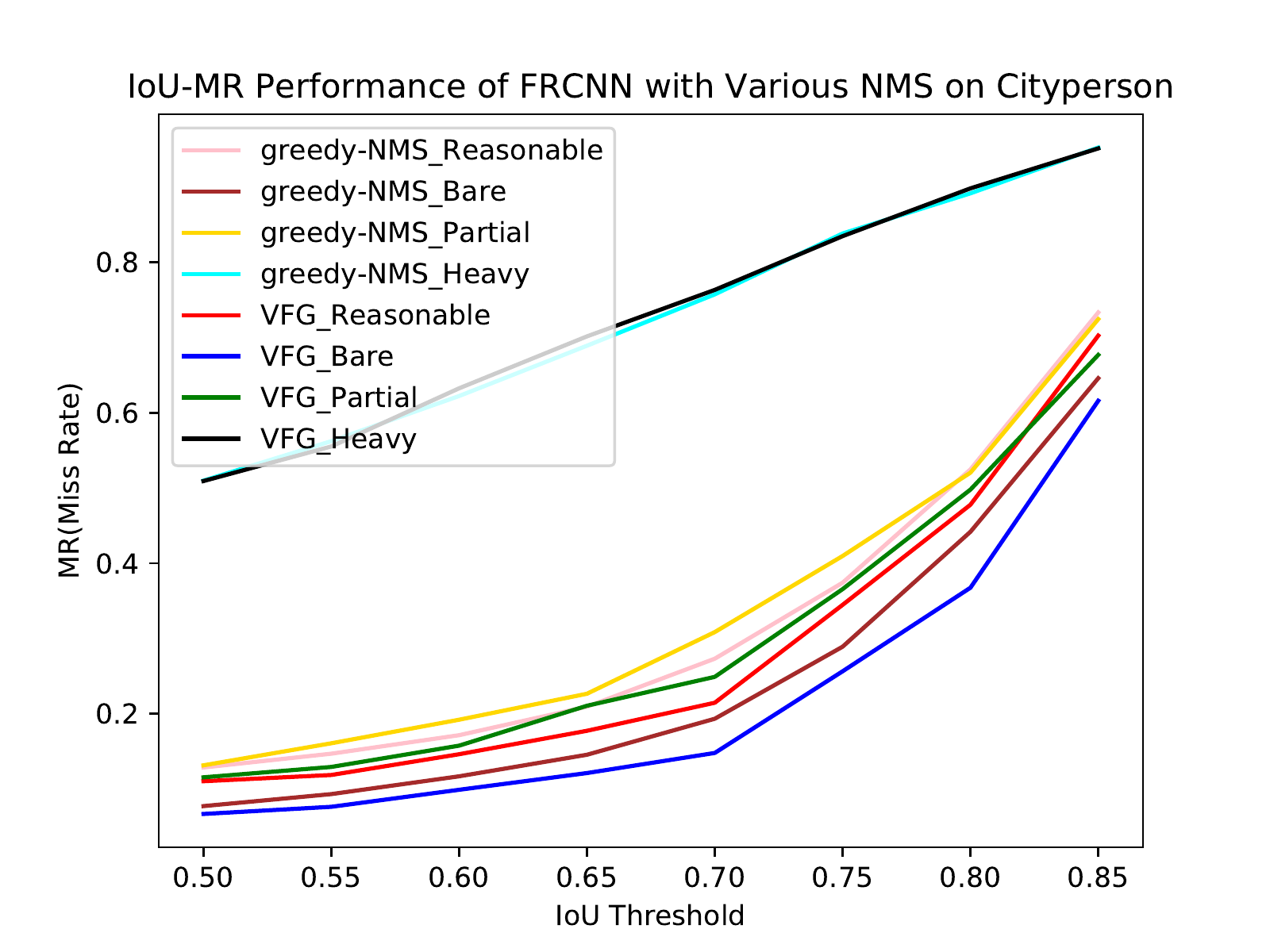}
			\caption{{IoU-MR on Cityperson}}
		\end{subfigure}
		\begin{subfigure}[t]{0.315\textwidth}
			\captionsetup{font={scriptsize}} 
			\centering
			\includegraphics[width=1.0\textwidth]{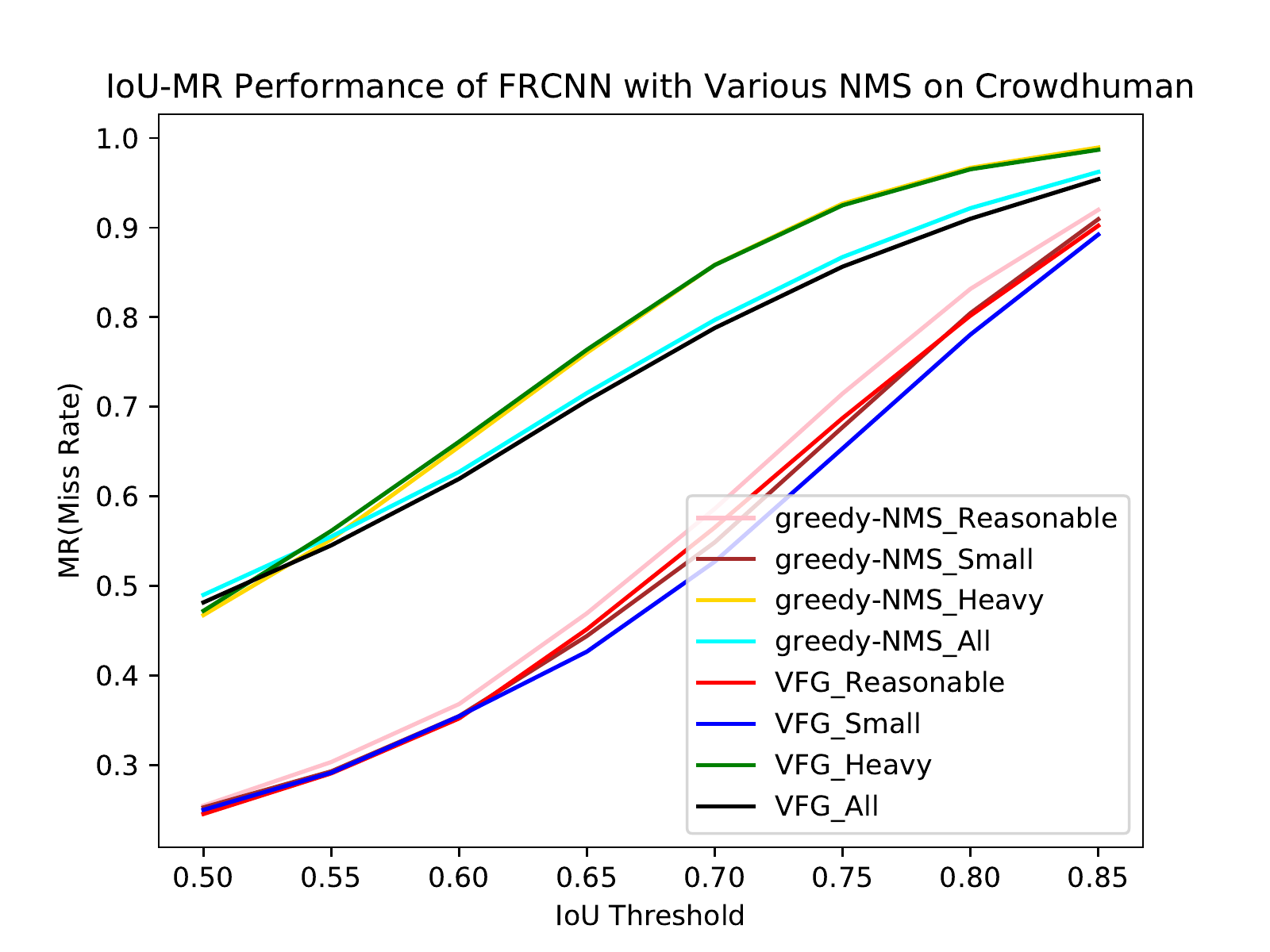}
			\caption{{IoU-MR on Crowdhuman}}
		\end{subfigure}
		\begin{subfigure}[t]{0.315\textwidth}
			\captionsetup{font={scriptsize}} 
			\centering
			\includegraphics[width=1.0\textwidth]{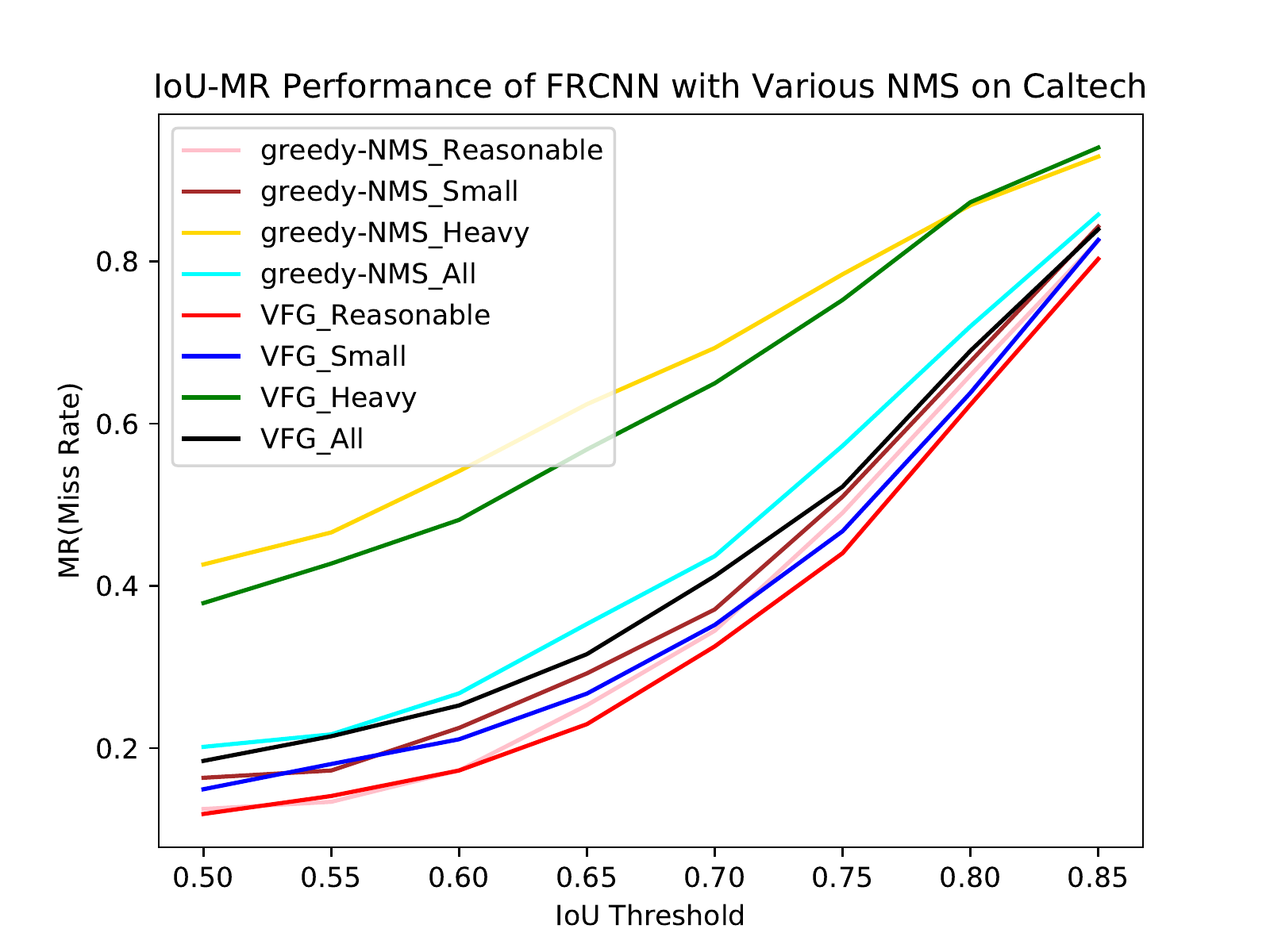}
			\caption{{IoU-MR on Caltech}}
		\end{subfigure}
		\caption{\small{
				\textbf{IoU Influence of MR Evaluation.} 
				With stricter IoU to calculate $\text{MR}^{-2}$, it shows VFG method is more robust.}}
		\label{fig:nms_iou_mMR}
		\vspace{-1.5\baselineskip}
	\end{figure*}
	\subsubsection{Caltech Dataset.}
	Whereas the work of Zhang \textit{et al.} \cite{zhang2016far} has provided refined annotations, the visible box annotations on Caltech are still noisy. 
	So we don't use this dataset for training.
	Instead, we evaluate models trained on Crowdhuman on the Caltech, verifying the generality of the proposed method. 
	In Table.\ref{table:ctbp}, compared to Faster R-CNN with greedy-NMS, the VFG method boosts mAP, $\text{AP}_{50}$ and recall with 3.2\%, 4.5\% and 8.4\%, reduces 0.6\%, 1.41\%, 4.76\% and 2.13\% $\text{MR}^{-2}$ in {\fontfamily{qcr}\selectfont Reasonable}, {\fontfamily{qcr}\selectfont Small}, {\fontfamily{qcr}\selectfont Heavy} and {\fontfamily{qcr}\selectfont All} subsets respectively. These results indicate generality of our VFG method. 
	\begin{figure*}[t]
		\centering
		\begin{subfigure}[t]{0.48\textwidth}
			\centering
			\includegraphics[width=1.0\textwidth]{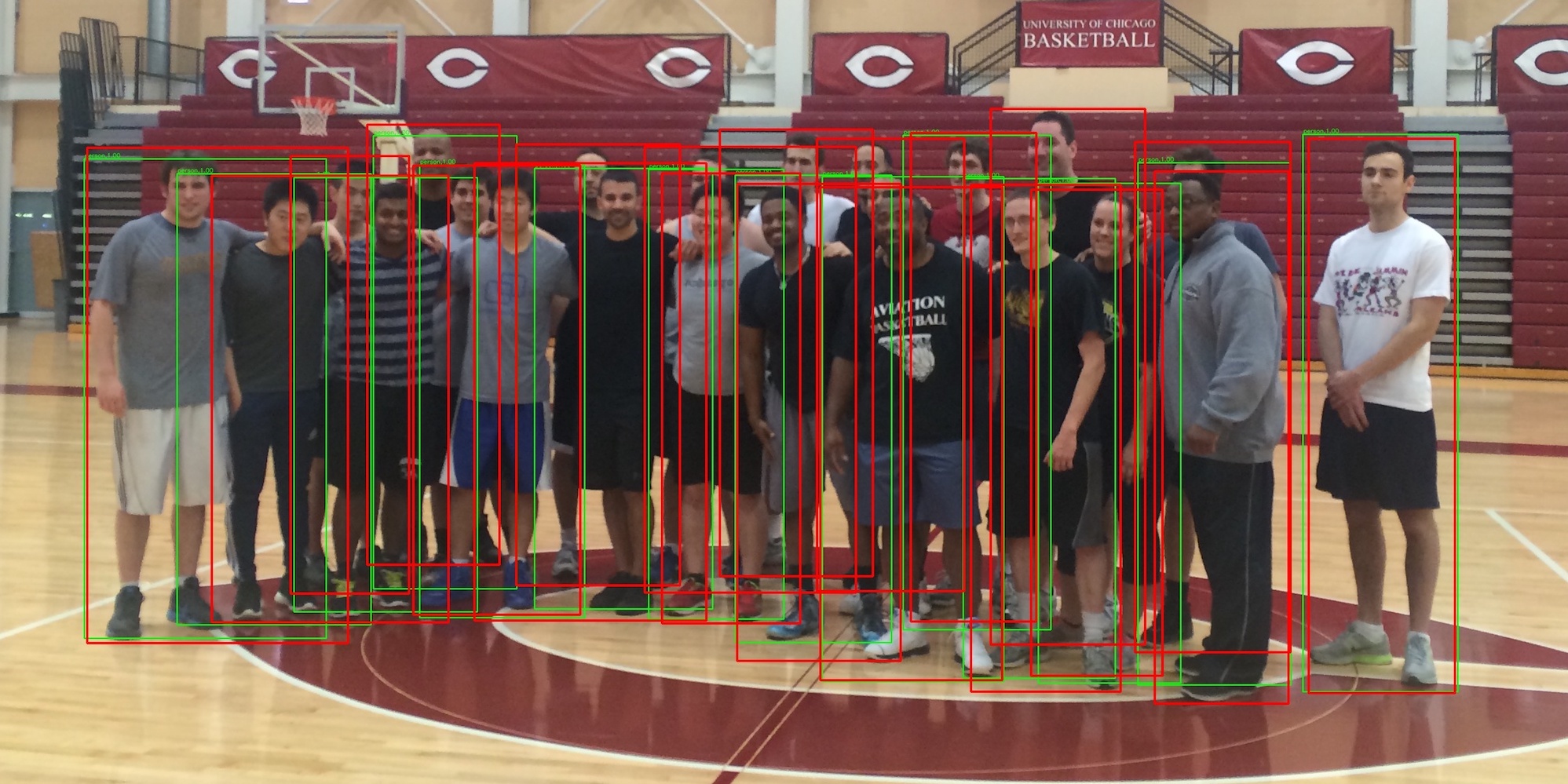}
		\end{subfigure}
		\begin{subfigure}[t]{0.48\textwidth}
			\centering
			\includegraphics[width=1.0\textwidth]{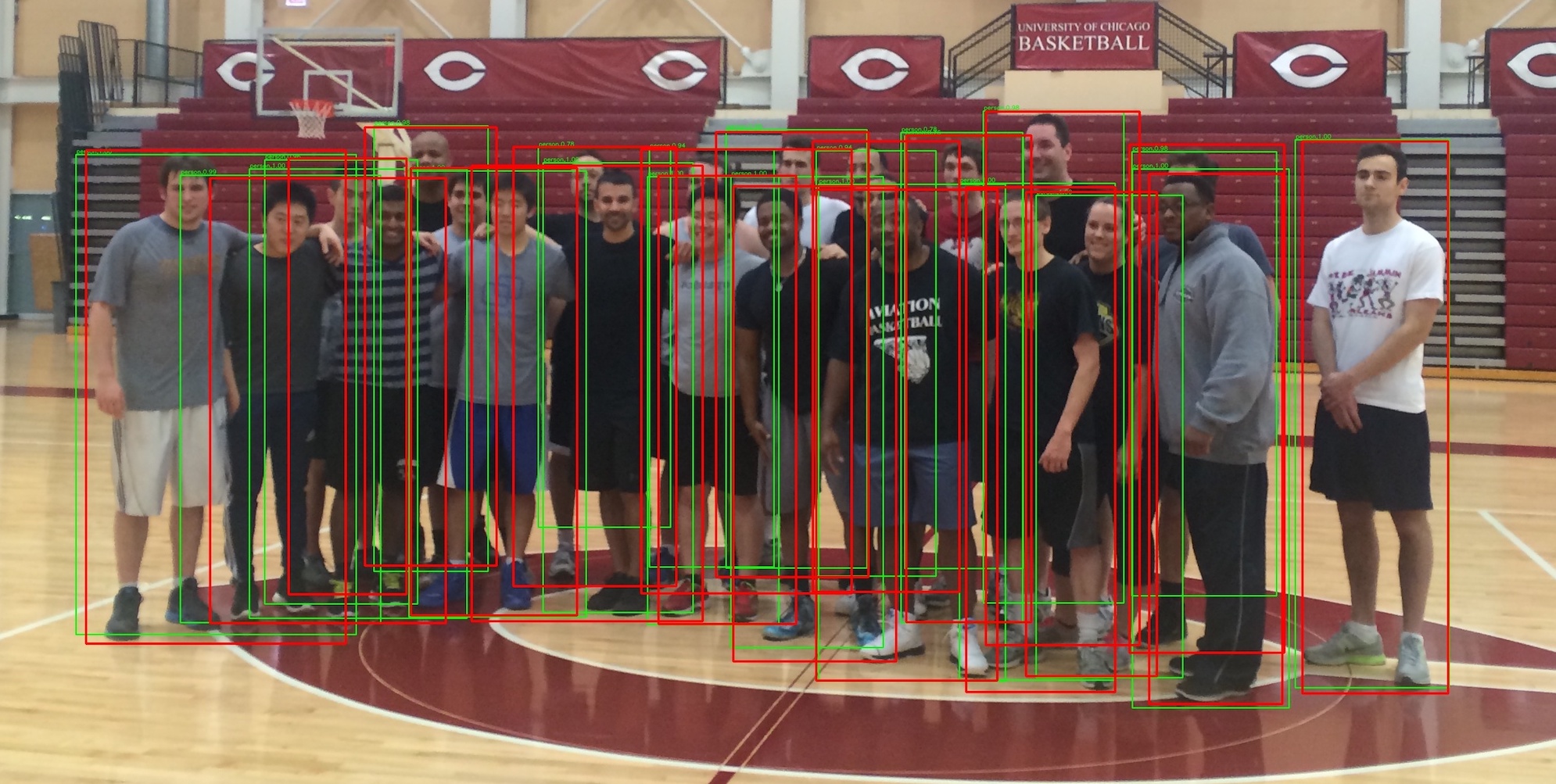}
		\end{subfigure}
		
		\begin{subfigure}[t]{0.48\textwidth}
			\centering
			\includegraphics[width=1.0\textwidth]{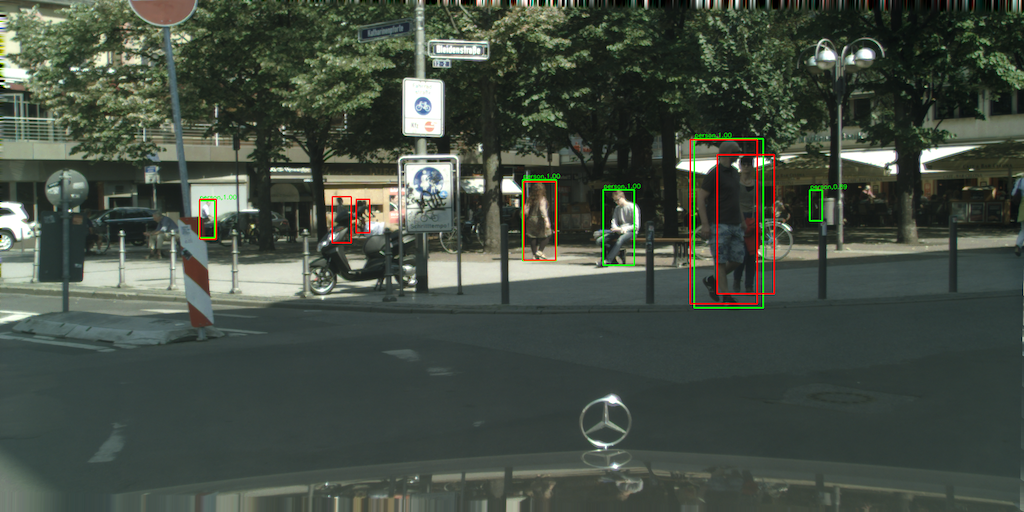}
		\end{subfigure}
		\begin{subfigure}[t]{0.48\textwidth}
			\centering
			\includegraphics[width=1.0\textwidth]{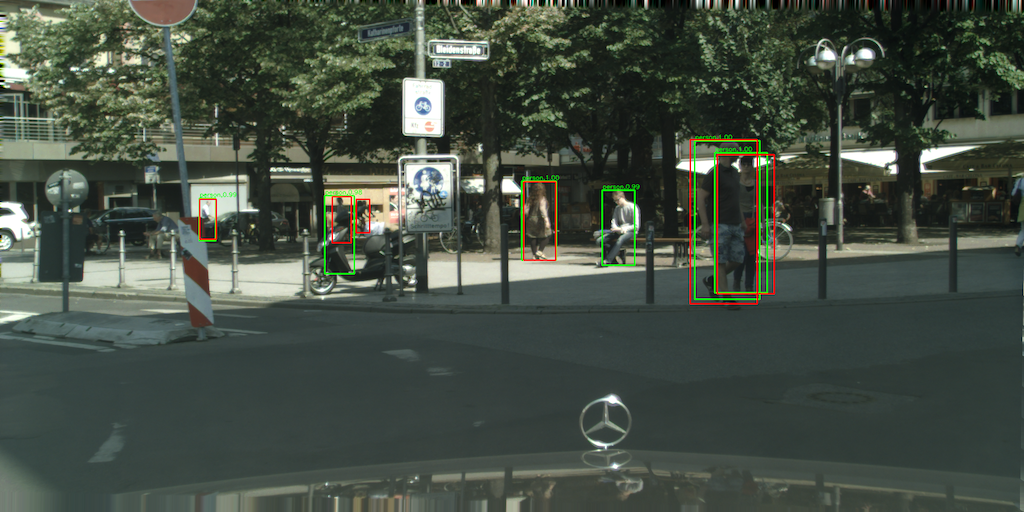}
		\end{subfigure}
		
		\begin{subfigure}[t]{0.48\textwidth}
			\centering
			\includegraphics[width=1.0\textwidth]{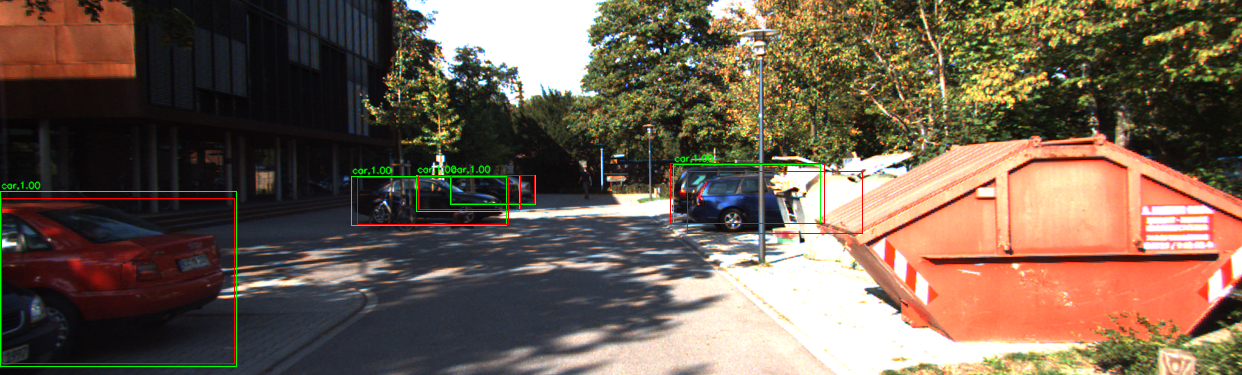}
			\caption{FRCNN+greedy-NMS}
		\end{subfigure}
		\begin{subfigure}[t]{0.48\textwidth}
			\centering
			\includegraphics[width=1.0\textwidth]{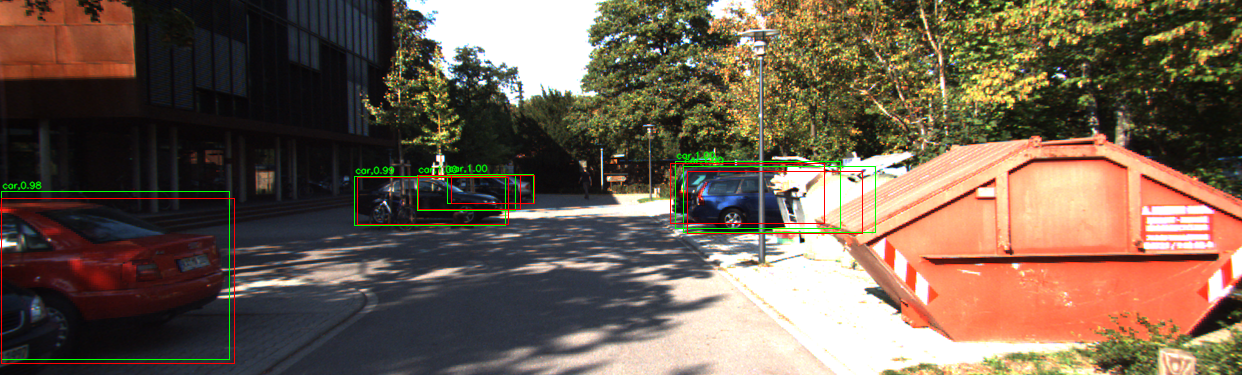}
			\caption{FRCNN+VFG}
		\end{subfigure}
		\caption{\small{
				\textbf{Visualization of Detection Result.} Red and green boxes mean ground-truth and predicted box respectively. The left (a) is the result of FRCNN with greedy-NMS method while right (b) is the result of our FRCNN with VFG method. The top, middle and bottom images are from Crowdhuman, Cityperson and KITTI datasets respectively.
		}}
		\label{fig:pro_recall}
		\vspace{-1.5\baselineskip}
	\end{figure*}
	\begin{table}[t]
		\centering
		\scriptsize
		\caption{\small{
				\textbf{Performances of full bounding box} on Caltech.
		}}
		\label{table:ctbp}
		\begin{tabular}{llllllll}
			\cmidrule{1-8}
			\multirow{2}{*}{method} & \multirow{2}{*}{mAP} & \multirow{2}{*}{$\text{AP}_{50}$}  & \multirow{2}{*}{Recall}  & \multicolumn{4}{c}{$\text{MR}^{-2}$}  \\ \cline{5-8} & & & & Reasonable & Small & Heavy & All  \\
			\toprule
			FRCNN+greedy & 26.4 & 53.5 & 57.0 & 12.49  & 16.35  & 42.64 &  46.47 \\
			FRCNN+soft & 27.2 & 53.6 & 58.7 & 12.54  & 16.49  & 42.19 & 46.68  \\
			FRCNN+VFG w/o VFG-NMS & 29.5 & 57.7 & 65.8 & 11.72  & 14.96  & 37.95 &  44.19 \\
			FRCNN+VFG & \textbf{29.6} & \textbf{58.0}  & \textbf{66.4} &  \textbf{11.89} & \textbf{14.94} & \textbf{37.88} & \textbf{44.34}  \\
			\cmidrule{1-8}
		\end{tabular}
		\vspace{-1.5\baselineskip}
	\end{table}
	\subsubsection{KITTI Dataset.}
\begin{figure}[h]
		\begin{minipage}[]{0.65\textwidth}
			\centering
			\scriptsize
			 \makeatletter\def\@captype{table}\makeatother
			 \caption{\small{\textbf{Performances of full bounding box} on KITTI Car.}}
			\vspace{-1\baselineskip}
			\begin{tabular}{llllll}
				\cmidrule{1-6}
				method & score & mAP & AP70 & AP80 & AP90 \\
				\toprule
				FRCNN+greedy & 0.05 & 69.8 & 86.4 & 75.9 & 29.5  \\
				FRCNN+VFG w/o VFG-NMS & 0.05 & 70.9 & 87.5 & 77.9 & 34.4  \\
				FRCNN+VFG & 0.05 & \textbf{71.7} & \textbf{87.6}  & \textbf{78.7} & \textbf{34.4} \\
				\cmidrule{1-6}
				FRCNN+greedy & 0.3 & 69.2 & 85.6 & 75.9 & 29.5 \\
				FRCNN+VFG w/o VFG-NMS & 0.3 & 69.8 & 85.9 & 77.1 & 34.0 \\
				FRCNN+VFG & 0.3 & \textbf{70.5} & \textbf{86.8}  & \textbf{78.0} & \textbf{34.4}  \\
				\cmidrule{1-6}
			\end{tabular}
			\label{table:cp}
		\end{minipage}
		\begin{minipage}[]{0.3\textwidth}
			\centering
			\scriptsize
			\includegraphics[width=1.\textwidth]{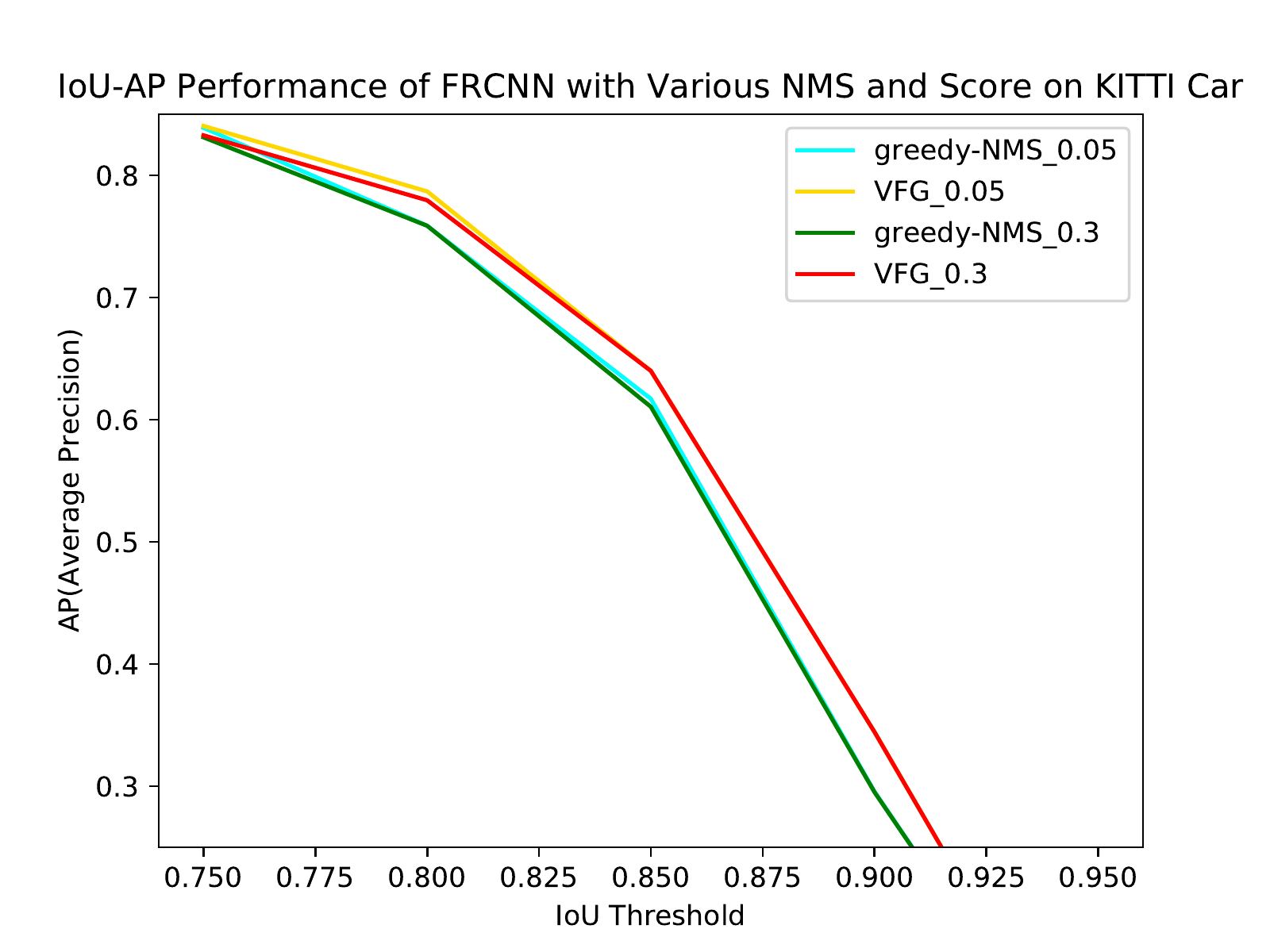}
			\vspace{-1.5\baselineskip}
			\caption{\small{IoU-AP Car Performance}}
			\label{fig:iouap}
		\end{minipage}
		\vspace{-1.5\baselineskip}
	\end{figure}

	For KITTI dataset, we choose the category of car to evaluate our proposed VFG method. Since the KITTI dataset provides full boxes only. In this experiment, we use Mask R-CNN pre-trained on COCO to produce the visible boxes of car and associate them with full boxes by Hungarian Algorithm.  As Table \ref{table:cp} shown, VFG method outperforms the FRCNN with greedy-NMS 1.9\% , 2.8\% and 4.9\% in mAP, AP80 and AP90 respectively with 0.05 detection score. To evaluate the performance further, we set the confidence threshold to 0.3 and get consistent result.
	%
	%
	
	\subsubsection{Ablation Study and Robustness Analysis.}
	To better investigate our VFG, we decompose its pipeline into two isolated flows: regression guided by visible feature and VFG-NMS. 
	We experiment ablations on these two aspects in the whole of above four datasets.
	When using visible feature guidance for regressor only (without VFG-NMS),
	the detector have improvements on mAP and $\text{MR}^{-2}$, especially with larger visible feature region over the whole instance area.
	If combining VFG-NMS, the final performances are boosted dramatically that remains more effective full boxes as shown in Fig.~\ref{fig:pro_recall}.
	By default, IoU threshold for evaluation $\text{MR}^{-2}$ is 0.5. 
	To comprehensively analyze the robustness of VFG, we plot curves of $\text{MR}^{-2}$ and mAP in accordance of higher IoU thresholds in Fig. \ref{fig:nms_iou_mMR} and Fig. \ref{fig:iouap} respectively.
	They show our VFG is more robust than greedy-NMS.
	It decreases $\text{MR}^{-2}$ by about 2\% in {\fontfamily{qcr}\selectfont Reasonable} subset over the stricter IoU threshold. 
	It also reflects that VFG can effectively prevent full boxes from being filtered out. 
	
	\subsection{Parts Association and Benchmark}
	To completely investigate the algorithm for post parts association, we do experiments with two tasks on Crowdhuman: (1) we associate body-head parts; (2) we decompose the pair output of detector and to associate the visible and full bounding boxes. 
	The experimental results build a simple and strong benchmark on Crowdhuman.
	
	First, we describe the evaluation method for the task of parts associations using an example, e.g, body-head.
	The pair of ground truth for body-head is indicated by $(B_{gt},H_{gt})$.
	The predicted pair of it by our proposed algorithm is $(B, H)$.
	The validation condition is $\frac{area(B_{gt} \cup B )}{area(B_{gt}\cap B)} \geq thresh_{B} \ \text{and} \  \frac{area(H_{gt} \cup H )}{area(H_{gt}\cap H)} \geq thresh_{H} $
	where $thresh_{B}$ is the threshold for body and $thresh_{H}$ for head to measure their relationship. When $(B, H)$ meets the above condition, it will match the ground truth. Otherwise it won't. 
	
\begin{figure}[t]
		\begin{minipage}[t]{0.5\linewidth}
			\centering
			\scriptsize
			\makeatletter\def\@captype{table}\makeatother
			\caption{\small{Body-Head Association}}
			\vspace{-1\baselineskip}
			\label{table:crowdhuman_body_head}
			\centering
			\begin{tabular}{llll}
				\cmidrule{1-4}
				Box-type & Score & Recall & Precision  \\
				\cmidrule{1-4}
				GT Box &  -- &  1.0 & 1.0    \\
				Pred Box & 0.05 & 73.83 & 64.58  \\
				Pred Box & 0.3 & 65.01 &  83.20 \\
				Pred Box & 0.7 &  56.99 &  92.14 \\
				\cmidrule{1-4}
			\end{tabular}
		\end{minipage}
		\begin{minipage}[t]{0.5\linewidth}
			\centering
			\scriptsize
			\makeatletter\def\@captype{table}\makeatother
			\caption{\small{Visible-full Body Association}}
			\vspace{-1\baselineskip}
			\label{table:crowdhuman_visible_full}
			\centering
			\begin{tabular}{llll}
				\cmidrule{1-4}
				Box-type & Score & Recall & Precision  \\
				\cmidrule{1-4}
				GT Box  & --  & 1.0 &  1.0   \\
				Pred Box & 0.05 &  87.23 &   61.58 \\
				Pred Box & 0.3 &  81.91 &  80.35 \\
				Pred Box & 0.7 & 73.45  & 90.33  \\
				\cmidrule{1-4}
			\end{tabular}
			\vspace{1.0\baselineskip}
		\end{minipage}
	
		\begin{minipage}[t]{1.\linewidth}
			\centering
			\begin{subfigure}[t]{0.48\textwidth}
				\centering
				\includegraphics[width=1.0\textwidth,height=3cm]{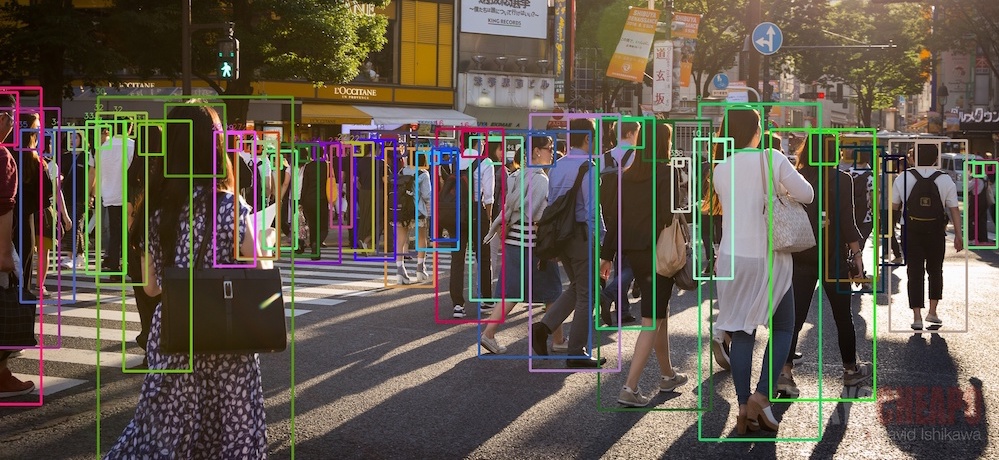}
				\vspace{-1.\baselineskip}
				\caption{Full Body-Head Association}
			\end{subfigure}
			\begin{subfigure}[t]{0.48\textwidth}
				\centering
				\includegraphics[width=1.0\textwidth, height=3cm]{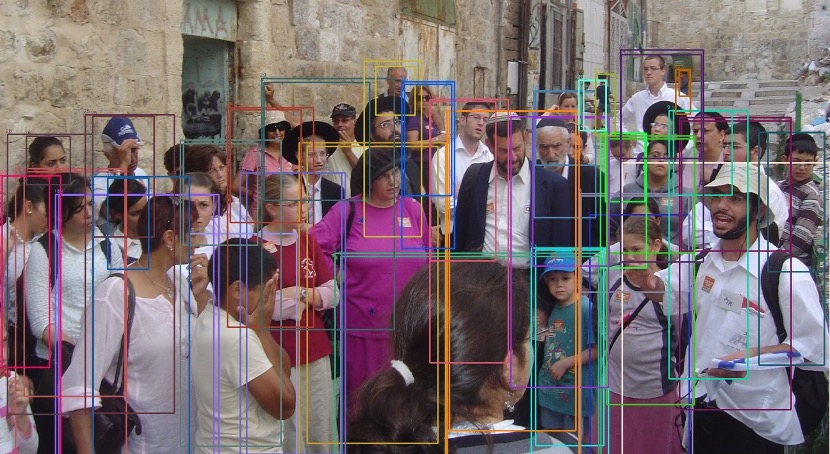}
				\vspace{-1.\baselineskip}
				\caption{Visible-Full Body Association}
			\end{subfigure}
			\vspace{0.\baselineskip}
			\caption{\textbf{Visualization of Parts Association Result.} Same color of body-parts box means belonging to the same person}
			\label{fig:bh_asso}
		   \vspace{-1.5\baselineskip}
		\end{minipage}
	\end{figure}

	 Table \ref{table:crowdhuman_body_head} and \ref{table:crowdhuman_visible_full} show the Hungarian algorithm can achieve the optimal solution with \textit{un-associated} ground-truth boxes on Crowdhuman. 
	The results of recall and precision are fully correct.
	This upper bound verify the correctness and effect of our matching algorithm. 
	In addition, we have trained the detectors for regressing out head, visible and full box respectively for performing association as shown in Fig \ref{fig:bh_asso}.
	In contrast, the predicted box (Pred box) output by deep learning model produces bad results due to the unsatisfied detected bounding boxes in crowd scene.
	As shown in Table \ref{table:crowdhuman_body_head} and \ref{table:crowdhuman_visible_full}, increasing the detection score threshold can make precision of association increase but recall decrease. This suggests that parts association largely depends on parts detection performance and we can conclude better detection could bring better association. 
	\section{Discussion}
	We have evaluated our proposed VFG on four popular datasets and show it works stably well in two-stage and one-stage detector.
	All the experimental results verify that VFG can stably promote 2$\sim$3\% in mAP and $\text{AP}_{50}$ and also be more effective for $\text{MR}^{-2}$ especially with stricter IoU.
	Moreover, we have benchmarked strongly for the task of parts association using Hungarian algorithm on Crowdhuman.
	Although all the ablations show the effect of VFG, they suggest that there is an interesting topic on reducing false positives further without loss of recall in the future research.
	\clearpage
	%
	%
	\bibliographystyle{splncs04}
	\bibliography{egbib}
\end{document}